\documentclass[letterpaper, 10 pt, conference]{ieeeconf}  

\IEEEoverridecommandlockouts                              

\usepackage{graphicx} 
\usepackage[]{booktabs}
\usepackage{cite}

\title{\LARGE \bf
Multi-Modal Trajectory Prediction of Surrounding Vehicles with \\ Maneuver based LSTMs   
}

\author{Nachiket Deo and Mohan M. Trivedi
\thanks{The authors are with the Laboratory for Intelligent and Safe Automobiles, University of California, San Diego, CA 92092, USA. 
        {\tt\small ndeo@ucsd.edu, mtrivedi@ucsd.edu}}%
}
\begin{document}

\maketitle
\thispagestyle{empty}
\pagestyle{empty}

\begin{abstract}
To safely and efficiently navigate through complex traffic scenarios, autonomous vehicles need to have the ability to predict the future motion of surrounding vehicles. Multiple interacting agents, the multi-modal nature of driver behavior, and the inherent uncertainty involved in the task make motion prediction of surrounding vehicles a challenging problem. In this paper, we present an LSTM model for interaction aware motion prediction of surrounding vehicles on freeways. Our model assigns confidence values to maneuvers being performed by vehicles and outputs a multi-modal distribution over future motion based on them. We compare our approach with the prior art for vehicle motion prediction on the publicly available NGSIM US-101 and I-80 datasets. Our results show an improvement in terms of RMS values of prediction error. We also present an ablative analysis of the components of our proposed model and analyze the predictions made by the model in complex traffic scenarios.
\end{abstract}

\section{Introduction}

An autonomous vehicle deployed in complex traffic needs to balance two factors: the safety of humans in and around it, and efficient motion without stalling traffic. The vehicle needs to have the ability to take initiative, such as, deciding when to change lanes, cross unsignalized intersections, or overtake another vehicle. This requires the autonomous vehicle to have some ability to reason about the future motion of surrounding vehicles. This can be seen in existing tactical path planning algorithms \cite{ulbrich,nilsson,dpdm}, all of which depend upon reliable estimation of future trajectories of surrounding vehicles.

Many approaches use motion models for predicting vehicle trajectories \cite{motion4,motion5,motion6,imm1}. However, motion models can be unreliable for longer prediction horizons, since vehicle trajectories tend to be highly non-linear due to the decisions made by the driver. This can be addressed by data-driven approaches to trajectory prediction \cite{soclstm,self,laug,vgmm}. These approaches formulate trajectory prediction as a regression problem by minimizing the error between predicted and true trajectories in a training dataset. A pitfall for regression based approaches is the inherent multi-modality of driver behavior. A human driver can make one of many decisions under the same traffic circumstances. For example, a driver approaching their leading vehicle at a faster speed could either slow down, or change lane and accelerate to overtake. Regression based approaches have a tendency to output the average of these multiple possibilities, since the average prediction minimizes the regression error. However, the average prediction may not be a good prediction. For instance, in the example scenario described above, the average prediction would be to stay in lane without deceleration. Thus, we need trajectory prediction models that address the multi-modal nature of predictions.

In this paper, we use \textit{maneuvers} for multi-modal trajectory prediction, by learning a model that assigns probabilities for different maneuver classes, and outputs maneuver specific predictions for each maneuver class. Following the success of Long-Short Term Memory (LSTM) networks in modeling non-linear temporal dependencies in sequence learning and generation tasks \cite{soclstm,graves,cho}, we propose an LSTM model for vehicle maneuver and trajectory prediction for the case of freeway traffic. It uses as input the track histories of the vehicle and its surrounding vehicles, and the lane structure of the freeway. It assigns confidence values to six maneuver classes and predicts a multi-modal distribution over future motion based on them. We train and evaluate our model using the NGSIM US-101\cite{ngsim1} and I-80 \cite{ngsim2} datasets of real vehicle trajectories collected on Californian multi-lane freeways.

\begin{figure}[t]
\centering
\includegraphics[width=\columnwidth]{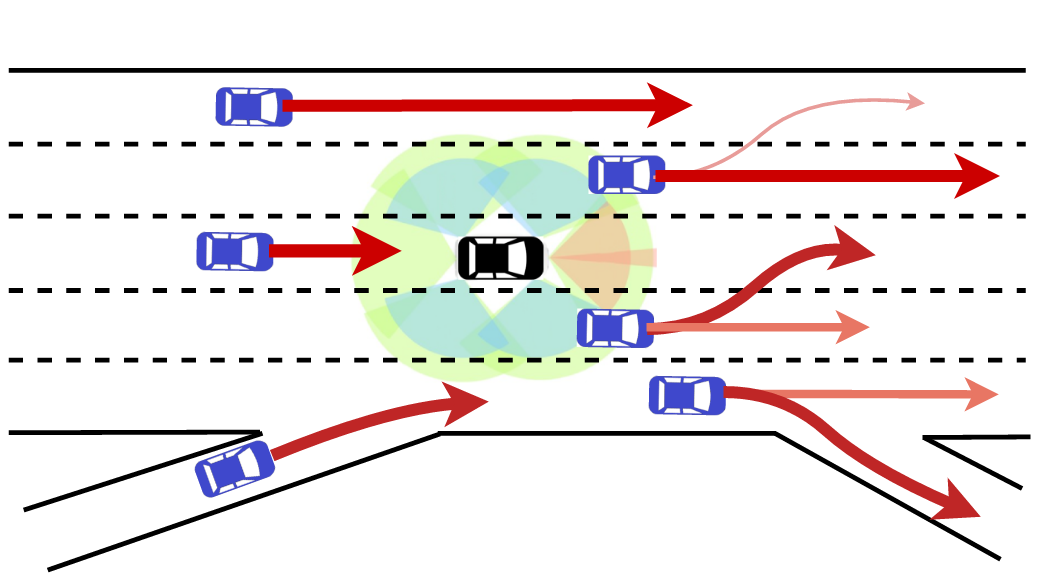}
\caption{An autonomous vehicle deployed in complex traffic (shown in the middle), needs to have the ability to predict the future motion of surrounding vehicles. Our proposed LSTM model allows for non-linear and multi-modal predictions of surrounding vehicle trajectories based on maneuver classes. It also assigns a probability to each mode and outputs uncertainty of prediction around each mode.}
\label{figurelabel}
\end{figure}

\section{Related Research}
\textbf{Maneuver based models:}
Classification of vehicle motion into maneuver classes has been extensively addressed in both advanced driver assistance systems as well as naturalistic drive studies\cite{jm,btm,ss2,aida,phil}. A comprehensive survey of maneuver-based models can be found in \cite{surv,ss1}. Of particular interest are works that use the recognized maneuvers to make better predictions of future trajectories \cite{houenou,schreier,tran,laug,gmrlat,self}. These approaches usually involve a maneuver recognition module for classifying maneuvers and maneuver specific trajectory prediction modules. Maneuver recognition modules are typically classifiers that use past positions and motion states of the vehicles and context cues as features. Heuristic based classifiers \cite{houenou}, Bayesian networks \cite{schreier}, hidden Markov models \cite{laug,self}, random forest classifiers \cite{gmrlat} and recurrent neural networks have been used for maneuver recognition. Trajectory prediction modules output the future locations of the vehicle given its maneuver class. Polynomial fitting \cite{houenou}, maneuver specific motion models \cite{schreier}, Gaussian processes \cite{tran,laug}, Gaussian mixture models \cite{self} have been used for trajectory prediction. Many approaches \cite{self,bahram,vi1,gmrlat} also take into consideration the interaction between vehicles for assigning maneuver classes and predicting trajectories. Hand crafted cost functions based on relative configurations of vehicles are used in \cite{self,bahram} to make optimal maneuver assignments for all surrounding vehicles. However, these approaches can be limited by how well the cost function is designed. Other works \cite{vi1,gmrlat} implicitly learn vehicle interaction from trajectory data of real traffic. Here we adopt the second approach due to the availability of large datasets of real freeway traffic \cite{ngsim1,ngsim2}.         

\textbf{Recurrent networks for motion prediction:}
Since motion prediction can be viewed as a sequence classification or sequence generation task, a number of LSTM based approaches have been proposed in recent times for maneuver
classification and trajectory prediction. Khosroshahi \textit{et al.} \cite{aida} and Phillips \textit{et al.} \cite{phil} use LSTMs to classify vehicle maneuvers at intersections. Kim \textit{et al.} \cite{kim} propose an LSTM that predicts the location of vehicles in an occupancy grid at intervals of 0.5s, 1s and 2s into the future. Contrary to this approach, our model outputs a continuous, multi-modal probability distribution of future locations of the vehicles up to a prediction horizon of 5s. Alahi \textit{et al.} \cite{soclstm} propose \textit{social LSTMs}, which jointly model and predict the motion of pedestrians in dense crowds through the use of a social pooling layer. However, vehicle motion on freeways has a lot more structure than pedestrians in crowds, which can be exploited to make better predictions. In particular, relative positions of vehicles can be succinctly described in terms of lane structure and direction of travel, and vehicle motion can be binned into maneuver classes, the knowledge of which can improve motion prediction. Lee \textit{et al.} \cite{lee} use an RNN encoder-decoder based conditional variational auto-encoder (CVAE) for trajectory prediction. Sampling the CVAE allows for multi-modal predictions. Contrarily, our model outputs the multi-modal distribution itself. Finally, Kuefler \textit{et al.} \cite{gail} use a gated recurrent unit (GRU) based policy using the behavior cloning and generative adversarial imitation learning paradigms to generate the acceleration and yaw-rate values of a bicycle model of vehicle motion. We compare our trajectory prediction results with those reported in \cite{gail}.

\section{Problem Formulation}
We formulate motion prediction as estimating the probability distribution of the future positions of a vehicle conditioned on it's track history and the track histories of vehicles around it, at each time instant $t$.
\subsection{Frame of reference}
\begin{figure}[t]
\centering
\includegraphics[width=\columnwidth]{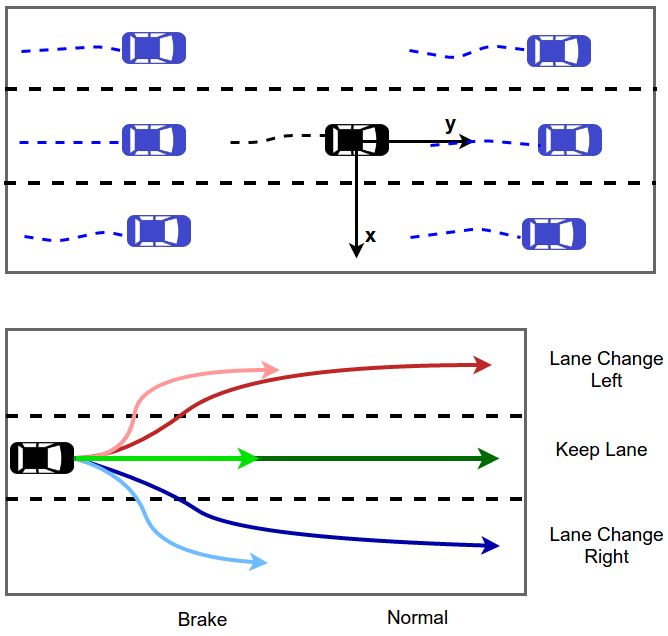}
\caption{\textbf{Top:} The co-ordinate system used for trajectory prediction. The vehicle being predicted is shown in black, neighboring vehicles considered are shown in blue. \textbf{Bottom:} Lateral and longitudinal maneuver classes}
\label{fig:form}
\end{figure}
 We use a stationary frame of reference, with the origin fixed at the vehicle being predicted at time $t$ as shown in Fig. \ref{fig:form}. The \textit{y-axis} points in the direction of motion of the freeway, and the \textit{x-axis} is the direction perpendicular to it. This  makes our model independent of how the vehicle tracks were obtained, and in particular, can be applied to the case of on-board sensors on an autonomous vehicle. This also makes the model independent of the curvature of the road, and can be applied anywhere on a freeway as long as an on-board lane estimation algorithm is available. 

\subsection{Inputs and outputs}
The input to our model is the tensor of track histories
$$\mathbf{X} = [\mathbf{x}^{(t-t_{h})},...,\mathbf{x}^{(t-1)}, \mathbf{x}^{(t)}]$$ 
where,
$$ \mathbf{x}^{(t)} = [x_{0}^{(t)},y_{0}^{(t)},x_{1}^{(t)},y_{1}^{(t)},...,x_{6}^{(t)},y_{6}^{(t)}]$$
are the $x$ and $y$ co-ordinates at time $t$ of the vehicle being predicted and six vehicles surrounding it as shown in Fig. \ref{fig:form}. We choose these six vehicles since they seem to have the most effect on a vehicle's motion. 

The output of the model is a probability distribution over 
$$\mathbf{Y} = [\mathbf{y}^{(t+1)},..., \mathbf{y}^{(t+t_{f})}]$$
where,
$$ \mathbf{y}^{(t)} = [x_{0}^{(t)},y_{0}^{(t)}]$$
are the future co-ordinates of the vehicle being predicted
\subsection{Probabilistic motion prediction}
Our model estimates the conditional distribution $\mbox{P}(\mathbf{Y}|\mathbf{X})$. In order to have the model produce multi-modal distributions, we expand it in terms of maneuvers $m_{i}$, giving:
\begin{equation}
\mbox{P}(\mathbf{Y}|\mathbf{X}) = \sum_{i}\mbox{P}_{\Theta}(\mathbf{Y}|m_{i},\mathbf{X})\mbox{P}(m_{i}|\mathbf{X})
\label{eq:prob}
\end{equation} 
where,
$$\Theta =[\Theta^{(t+1)},..., \Theta^{(t+t_{f})}]$$
are the parameters of a bivariate Gaussian distribution at each time step in the future, corresponding to the means and variances of future locations. 

\subsection{Maneuver classes}
We consider three lateral and two longitudinal maneuver classes as shown in Fig. \ref{fig:form}. The lateral maneuvers consist of left and right lane changes and a lane keeping maneuver. Since lane changes involve preparation and stabilization, we define a vehicle to be in a lane changing state for $\pm$ 4s w.r.t. the actual cross-over.
The longitudinal maneuvers are split into normal driving and braking. We define a vehicle to be performing a braking maneuver if it's average speed over the prediction horizon is less than 0.8 times its speed at the time of prediction. We define our maneuvers in this manner since these maneuver classes are communicated by vehicles to each other through turn signals and brake lights, which will be included as a cue in future work.

\section{Model}
\begin{figure}[t]
\centering
\includegraphics[width=\columnwidth]{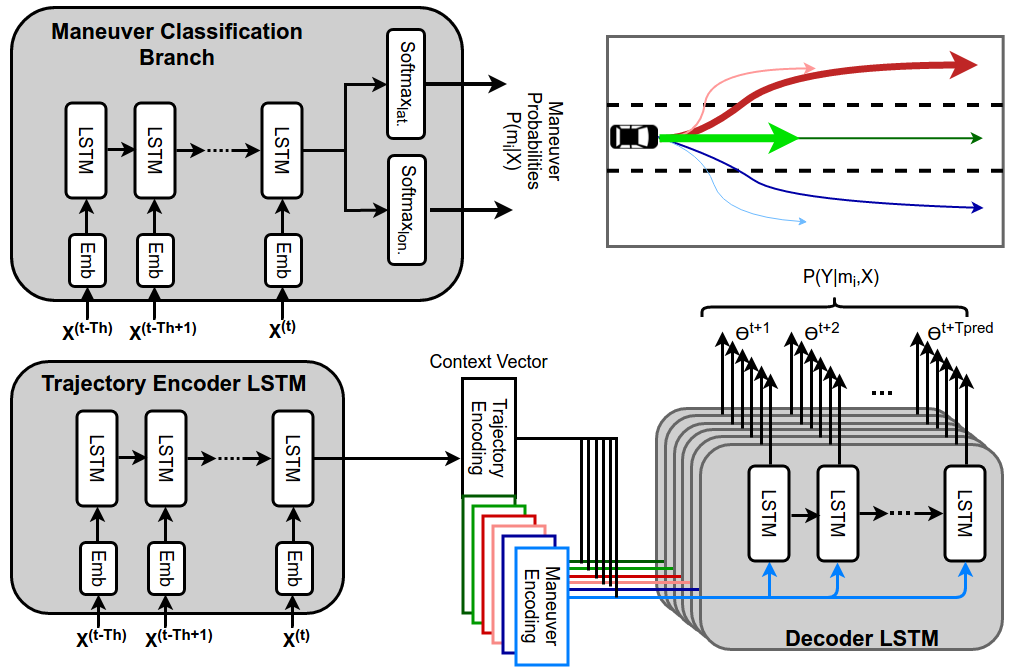}
\caption{\textbf{Proposed Model:} The trajectory encoder LSTM encodes the track histories and relative positions of the vehicle being predicted and its adjacent vehicles in a context vector. The context vector is appended with maneuver encodings of the lateral and longitudinal maneuver classes. The decoder LSTM generates maneuver specific future distributions of vehicle positions at each time step, and the maneuver classification branch assigns maneuver probabilities}
\label{fig:sys}
\end{figure}
\subsection{LSTM encoder-decoder} 
Our proposed model is shown in Fig. \ref{fig:sys}. We use an encoder-decoder framework\cite{cho}. The trajectory encoder LSTM takes as input the frame by frame past locations of the predicted vehicle and its six adjacent vehicles for the past $t_{h}$ frames. The hidden state vector of the encoder LSTM is updated at each time step based on the hidden state at the previous time step and the input frame of vehicle locations at the current time step. The final state of the trajectory encoder LSTM can be expected to encode information about the track histories and relative positions of the 7 vehicles. This context vector is then used by the decoder LSTM as input. At each time step, for $t_f$ frames into the future, the decoder LSTM state is updated based on the encoded context vector and LSTM state at the previous instant. The decoder outputs at each time step, a 5-D vector $\Theta^{(t)}$ corresponding to the parameters of a bivariate Gaussian distribution, giving the distribution of the future locations of the predicted vehicle at that time instant, conditioned on the track histories.     

\subsection{Maneuver dependent predictions}
The encoder-decoder model described in the previous section outputs a uni-modal maneuver-independent trajectory distribution. In order to have the decoder generate a multi-modal trajectory distribution based on the six maneuver classes defined, we append the encoder context vector with a one-hot vector corresponding to the lateral maneuver class and a one-hot vector corresponding to the longitudinal maneuver class. The added maneuver context allows the decoder LSTM to generate maneuver specific probability distributions $\mbox{P}_{\Theta}(\mathbf{Y}|m_{i},\mathbf{X})$ as given in Eq. \ref{eq:prob}. To obtain the conditional probabilities $\mbox{P}(m_{i}|\mathbf{X})$ for each maneuver class given track histories, we train the maneuver classification branch of the model shown in Fig. \ref{fig:sys}. The maneuver classification LSTM has the same inputs as the trajectory encoder LSTM. It has two output softmax layers for predicting the probabilities of lateral and longitudinal maneuver classes. Assuming the lateral and longitudinal maneuver classes to be conditionally independent given the track history, we obtain $\mbox{P}(m_{i}|\mathbf{X})$ by taking the product of the corresponding lateral and longitudinal maneuver probabilities.    

\subsection{Implementation details}
We use LSTMs with 128 units for the encoder, decoder and maneuver classification branch. The input vectors $X^{(t)}$ are embedded using a 64 unit fully-connected layer with leaky ReLU activation with $\alpha$=0.1, prior to being input to the LSTM layer. Although the trajectory encoder-decoder and maneuver classification models are used in tandem during test time, we train the models separately. The trajectory encoder-decoder is trained to minimize the negative log likelihood loss for the ground truth future locations of vehicles under the predicted trajectory distribution. The context vector is appended with the ground truth values of the maneuver classes for each training sample. The maneuver classification model is trained to minimize the the sum of cross-entropy losses of the predicted and ground truth lateral and longitudinal maneuver classes. Both models are trained using Adam \cite{adam} with a learning rate of 0.001.  The models are implemented using Keras \cite{keras}.

\section{Experimental Evaluation}
\begin{figure}[t]
\centering
\includegraphics[width=\columnwidth]{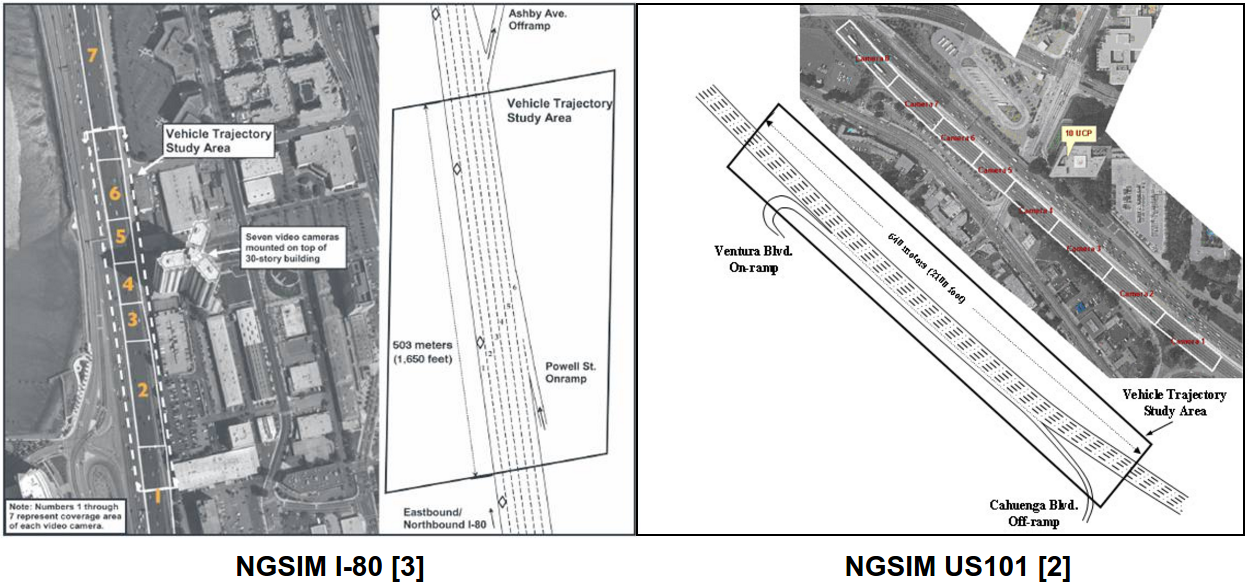}
\caption{\textbf{Dataset:} Layouts and top-down views of the sites used for collecting the NGSIM US-101 \cite{ngsim1} and NGSIM I-80 \cite{ngsim2} datasets used for evaluation. The datasets consists of real trajectories of vehicles on multi-lane freeways with entry and exit ramps, at varying traffic densities.}
\label{figurelabel}
\end{figure}

\subsection{Dataset}
We use the publicly available NGSIM US-101 \cite{ngsim1} and I-80 \cite{ngsim2} datasets for our experiments. Each dataset consists of trajectories of real freeway traffic captured at 10 Hz over a time span of 45 minutes. Each dataset consists of 15 min segments of mild, moderate and congested traffic conditions. The dataset provides the co-ordinates of vehicles projected to a local co-ordinate system, as defined in Section IIIA. We split the datasets into train and test sets. A fourth of the trajectories from each of the 3 subsets of the US-101 and I-80 datasets are used in the test set. We split the trajectories into segments of 8 s, where we use 3 s of track history and a 5 s prediction horizon. These 8 s segments are sampled at the dataset sampling rate of 10Hz. However we downsample each segment by a factor of 2 before feeding them to the LSTMs, to reduce the model complexity.      

\subsection{Models compared}

We report results in terms of RMS values of prediction error over a prediction horizon of 5 seconds as done in \cite{gail}. The following models are compared
\begin{itemize}
\item \textit{Constant Velocity (CV)}: We use a constant velocity Kalman filter as our simplest baseline
\item \textit{C-VGMM + VIM:} We use maneuver based variational Gaussian mixture models with a Markov random field based vehicle interaction module described in \cite{self} as our second baseline. We modify the model to use the maneuver classes described in this work to allow for a fair comparison
\item \textit{GAIL-GRU:} We consider the GRU model based on generative adversarial imitation learning described in \cite{gail}. Since the same datasets have been used in both works, we use the results reported by the authors in the original article
\item \textit{Maneuver-LSTM (M-LSTM) :} We finally consider the model proposed in this paper. Since each of the baselines makes a unimodal prediction, to allow for a fair comparison, we use the prediction corresponding to the maneuver with the highest probability as given by our proposed model  
\end{itemize}
\subsection{Results}
\begin{table}[t]
\centering
\caption{RMS values of prediction error}
\label{tab:results}
\begin{tabular}{@{}ccccc@{}}
\toprule
\begin{tabular}[c]{@{}c@{}}Prediction\\ horizon\\ (s)\end{tabular} & CV   & \begin{tabular}[c]{@{}c@{}}C-VGMM \\ + VIM\\ \cite{self}\end{tabular} & \begin{tabular}[c]{@{}c@{}}GAIL-GRU\\ \cite{gail}\end{tabular} & \begin{tabular}[c]{@{}c@{}}M-LSTM\\ (this work)\end{tabular} \\ \midrule
1                                                                  & 0.73 & 0.66                                                           & 0.69                                                    & \textbf{0.58}                                                \\
2                                                                  & 1.78 & 1.56                                                           & 1.51                                                    & \textbf{1.26}                                                \\
3                                                                  & 3.13 & 2.75                                                           & 2.55                                                    & \textbf{2.12}                                                \\
4                                                                  & 4.78 & 4.24                                                           & 3.65                                                    & \textbf{3.24}                                                \\
5                                                                  & 6.68 & 5.99                                                           & 4.71                                                    & \textbf{4.66}                                                \\ \bottomrule
\end{tabular}
\end{table}

\begin{figure}[t]
\centering
\includegraphics[width=\columnwidth]{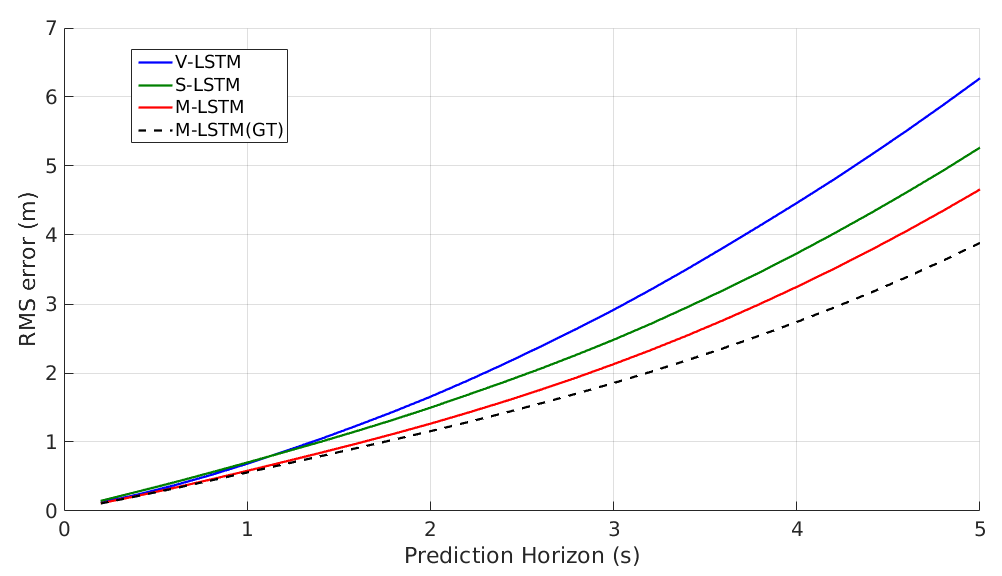}
\caption{ \textbf{Ablative analysis of components of the proposed model:} The RMS values of prediction error show the significance of modeling the track histories of adjacent vehicles in the trajectory encoder, and using the maneuver recognition model}
\label{fig:abl}
\end{figure}

Table \ref{tab:results} shows the RMS values of prediction error for the models being compared. We note that the proposed M-LSTM and the GAIL model from \cite{gail} considerably outperform the CV baseline and the C-VGMM + VIM model from \cite{self}, which suggests the superiority of recurrent neural networks in modeling non-linear motion of vehicles. In particular, the reduction in RMS values becomes more pronounced for longer prediction intervals. We also note that the M-LSTM achieves lower prediction error as compared to the GAIL model for all prediction intervals. Based on the trend of the error values, we see that the GAIL model seems to be catching up with the M-LSTM as the prediction horizon increases. However, we need to account for the fact that the GAIL trajectories in \cite{gail} were generated by running the policy one vehicle at a time, while all surrounding vehicles move according to the ground-truth of the NGSIM dataset. Thus, the model has access to the true trajectories of adjacent vehicles over the prediction horizon.   \begin{figure*}[t]
\centering
\includegraphics[width=\textwidth]{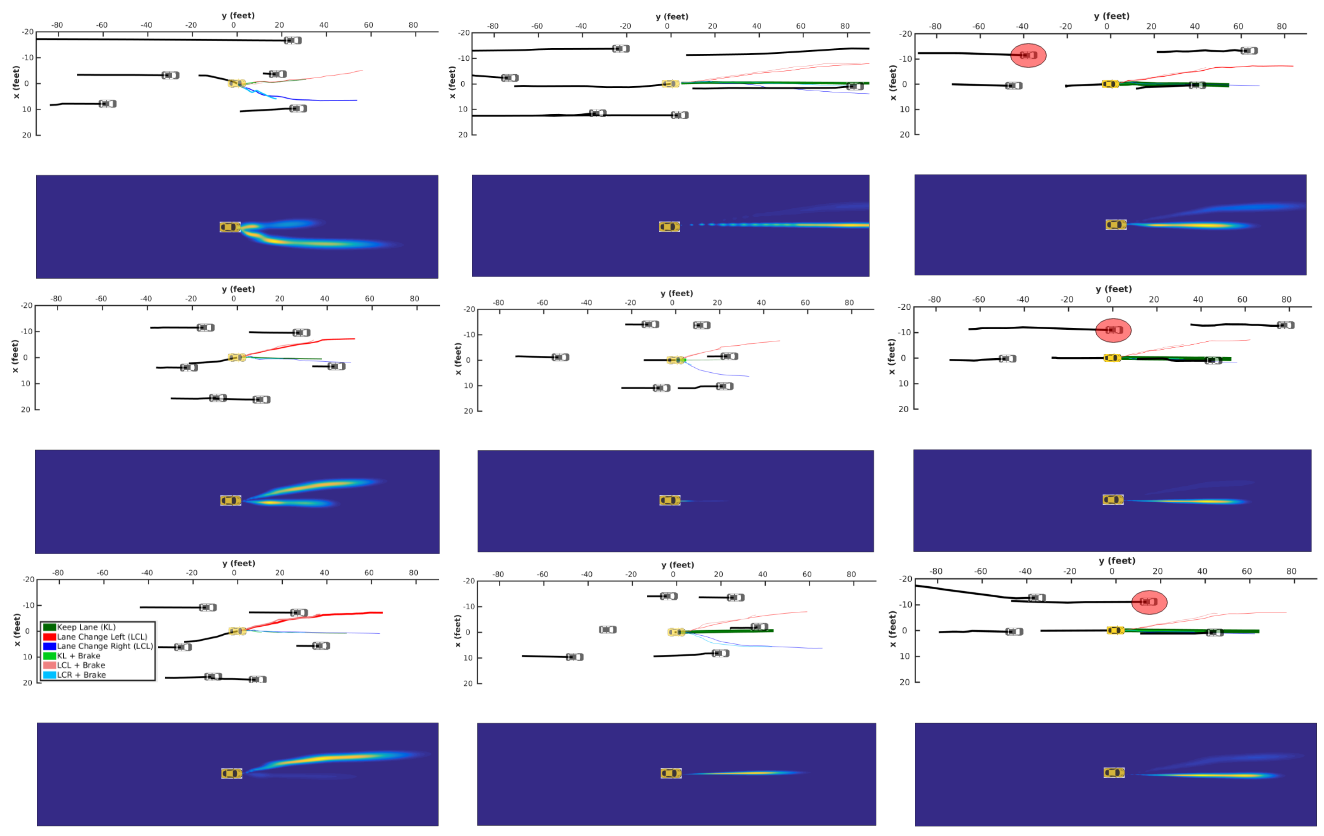}
\caption{ \textbf{Analysis of predictions:} (a) Multimodal predictions, (b) Effect of leading vehicle, (c) Effect of adjacent vehicle }
\label{fig:ex}
\end{figure*} 

\subsection{Ablative Analysis}

We conduct an ablative analysis of our model's components to study their relative significance for motion prediction. 
In particular, we seek to test the significance of using track histories of adjacent vehicles and of using the maneuver classification branch. We compare the RMS values of prediction error for the following system settings:
\begin{itemize}
\item \textit{Vanilla LSTM (V-LSTM):} This simply uses the predicted vehicle's track history in the encoder LSTM
\item \textit{Surround LSTM (S-LSTM):} This additionally considers the adjacent vehicle track histories in the encoder LSTM
\item \textit{Surround LSTM + Maneuver recognition (M-LSTM):} This considers the complete model proposed in the paper
\item \textit{Surround LSTM with ground truth maneuvers(M-LSTM (GT)):} Finally, we also consider the M-LSTM with ground truth values of maneuver classes, to gauge the potential improvement in trajectory prediction with improved maneuver recognition
\end{itemize}

Fig \ref{fig:abl} shows the RMS values of prediction error for the 4 system settings considered. We observe that the S-LSTM outperforms the Vanilla LSTM model, suggesting that motion of adjacent vehicles is a significant cue for predicting the future motion of vehicles. The M-LSTM leads to further improvement in prediction accuracy, suggesting the usefulness of maneuver classification prior to motion prediction. Both effects seem to become more pronounced for longer prediction intervals. Additionally, we note from the RMSE values of M-LSTM(GT) that considerable further improvement could have been achieved if maneuver classification was more accurate.

\subsection{Qualitative analysis of predictions}

In this section we qualitatively analyze the predictions made by our model to gain insights into its behavior in various traffic configurations. Figure \ref{fig:ex} shows six different scenarios of traffic. Each figure shows a plot of track histories over the past 3 seconds and the mean predicted trajectories over the next 5 seconds for each maneuver class. The thickness of the plots of the predicted trajectories is proportional to the probabilities assigned to each maneuver class. Additionally, each figure shows a heat map of the complete predicted distribution. 

Fig. 5(a) illustrates the multi-modal nature of the prediction made by the model for vehicles about to change lanes. The predicted distribution has a mode corresponding to the respective lane change, as well as the \textit{keep lane} maneuver. The model becomes more and more confident in the lane change further into the maneuver. We note that the model predicts the vehicle to merge into the target lane for the lane change maneuvers illustrating the ability of the LSTM to model the non-linear nature of vehicle motion.

Fig 5(b) shows the effect of the leading vehicle on the predictions made by the model. The first example shows an example of free flowing traffic, where the predicted vehicle and the leading vehicle are moving at approximately the same speed. In the second example, we note from the track histories that the leading vehicle is slowing down compared to the predicted vehicle. We see that the model predicts the vehicle to brake, although it's current motion suggests otherwise. Conversely, in the third example, we see that the vehicle being predicted is almost stationary, while the leading vehicle is beginning to move. The model predicts the vehicle to accelerate, as is expected in stop-and-go traffic.

Fig 5(c) shows the effect of vehicles in the adjacent lane on the model's predictions. The three examples show the same scenario separated by 0.5 sec, with the vehicle being predicted is in the rightmost lane. We note that in all three cases shown, the model assigns a high probability to the vehicle keeping lane. However it also assigns a small probability for the vehicle to change to the left lane. We note that the probability to change to the left lane is affected by the circled vehicle shown in the plots. When the circled vehicle is far behind, the model assigns a high probability to the lane change. When the vehicle is right next to the vehicle being predicted, the lane change probability drops. When the vehicle passes and the lane opens up again, the probability for lane change increases again.

\section{Conclusions}
A novel LSTM based interaction aware model for vehicle motion prediction was presented in this paper, capable of making multi-modal trajectory predictions based on maneuver classes. The model was shown to achieve lower prediction error on two large datasets of real freeway vehicle trajectories, compared to two existing state of the art approaches from literature, demonstrating the viability of the approach. Additionally, an ablative analysis of the system showed the significance of modeling the motion of adjacent vehicles for predicting the future motion of a given vehicle, and detecting and exploiting common maneuvers of vehicles for future motion prediction. 

\section{Acknowledgement}
We gratefully acknowledge the support of our colleagues and sponsors and the comments of the anonymous reviewers.


%
%





\end{document}